\documentclass[11pt]{article}
\usepackage{coling2020}
\usepackage{times}
\usepackage{url}
\usepackage{latexsym}

\usepackage{subcaption}
\usepackage{graphicx}
\usepackage{booktabs}
\usepackage{multirow}
\usepackage{makecell}
\usepackage{xspace}

\usepackage{xcolor}

\usepackage[hidelinks]{hyperref}

\usepackage{colortbl}
\newcommand*{\belowrulesepcolor}[1]{%
  \noalign{%
    \kern-\belowrulesep
    \begingroup
      \color{#1}%
      \hrule height\belowrulesep
    \endgroup
  }%
}
\newcommand*{\aboverulesepcolor}[1]{%
  \noalign{%
    \begingroup
      \color{#1}%
      \hrule height\aboverulesep
    \endgroup
    \kern-\aboverulesep
  }%
}

\newcommand{\TransQuest}{TransQuest\xspace}

\colingfinalcopy % Uncomment this line for the final submission

\title{\TransQuest: Translation Quality Estimation with \\ Cross-lingual Transformers

}

\author{Tharindu Ranasinghe$^\diamondsuit$, \textbf{Constantin Or\u{a}san$^\heartsuit$ and Ruslan Mitkov$^\diamondsuit$} \\
 $^\diamondsuit$Research Group in Computational Linguistics, University of Wolverhampton, UK \\
 $^\heartsuit$Centre for Translation Studies, University of Surrey, UK \\
 {\tt \{t.d.ranasinghehettiarachchige, r.mitkov\}@wlv.ac.uk} \\
 {\tt  c.orasan@surrey.ac.uk} }

\date{}

\begin{document}
\maketitle

\begin{abstract}

Recent years have seen big advances in the field of sentence-level quality estimation (QE), largely as a result of using neural-based architectures. However, the majority of these methods work only on the language pair they are trained on and need retraining for new language pairs. This process can prove difficult from a technical point of view and is usually computationally expensive. In this paper we propose a simple QE framework based on cross-lingual transformers, and we use it to implement and evaluate two different neural architectures. Our evaluation shows that the proposed methods achieve state-of-the-art results outperforming current open-source quality estimation frameworks when trained on datasets from WMT. In addition, the framework proves very useful in transfer learning settings, especially when dealing with low-resourced languages, allowing us to obtain very competitive results.
 
\end{abstract}

\section{Introduction}
\label{sec:intro}

\blfootnote{

    \hspace{-0.5cm}  % space normally used by the marker
    This work is licensed under a Creative Commons 
    Attribution 4.0 International Licence.
    Licence details:
    \url{http://creativecommons.org/licenses/by/4.0/}.
}

The goal of quality estimation (QE) is to evaluate the quality of a translation without having access to a reference translation \cite{specia-etal-2018-findings}.
High-accuracy QE that can be easily deployed for a number of language pairs is the missing piece in many commercial translation workflows as they have numerous potential uses. They can be employed to select the best translation when several translation engines are available or can inform the end user about the reliability of automatically translated content. In addition, QE systems can be used to decide whether a translation can be published as it is in a given context, or whether it requires human post-editing before publishing or even translation from scratch by a human  \cite{kepler-etal-2019-openkiwi}. The estimation of translation quality can be done at different levels: document level, sentence level and word/phrase level \cite{ive-etal-2018-deepquest}. In this research we focus on sentence-level quality estimation.   

As we discuss in Section \ref{sec:related}, at present neural-based QE methods constitute the state of the art in quality estimation. However, these approaches are based on complex neural networks and require resource-intensive training. This resource-intensive nature of these deep-learning-based frameworks makes it expensive to have QE systems that work for several languages at the same time. Furthermore, these architectures require a large number of annotated instances for training, making the quality estimation task very difficult for low-resource language pairs.

In this paper we propose \textit{\TransQuest}, a framework for sentence-level machine translation quality estimation which solves the aforementioned problems, whilst obtaining competitive results.
The motivation behind this research is to propose a simple architecture which can be easily trained with different types of inputs (i.e. different language pairs or language from different domains) and can be used for transfer learning in settings where there is not enough training data. 
We show that \textit{\TransQuest} outperforms current open-source quality estimation frameworks and compares favourably to winning solutions submitted to recent shared tasks on 15 different language pairs on different aspects of quality estimation. In fact, a tuned version of TransQuest was declared the winner for all 8 tasks of the direct assessment sentence level QE shared task organised at WMT 2020 (for more details see \cite{ranasinghe-etal-2020-transquest} and Section \ref{subsec:supervised}. The main contributions of this paper are the following: 

\begin{enumerate}
\item We introduce \textit{\TransQuest}, an open-source framework, and use it to implement two neural network architectures that outperform current state-of-the-art quality estimation methods in two different aspects of sentence-level quality estimation. 

\item To the best of our knowledge this is the first neural-based method which develops a model capable of providing quality estimation for more than one language pair. In this way we address the problem of high costs required to maintain a multi-language-pair QE environment.

\item We tackle the problem of quality estimation in low-resource language pairs by showing that even with a small number of annotated training instances, \textit{\TransQuest} with transfer learning can outperform current state-of-the-art quality estimation methods in low-resource language pairs.

\item We provide important resources to the community: the code as an open-source framework, as well as the \textit{\TransQuest} model zoo – a collection of pre-trained quality estimation models which have been trained on 15 different language pairs and different aspects of quality estimation – will be freely available to the community\footnote{The public GitHub repository is available on \url{https://github.com/tharindudr/transquest} and the official documentation is available on \url{https://tharindudr.github.io/TransQuest}.}.
\end{enumerate}

The remainder of the paper is structured as follows. We first present a brief overview of related work in order to define the context of our work. In Section \ref{sec:methodology} we present the \TransQuest framework and the methodology employed to train it. The datasets used to train it are presented in Section \ref{sec:datasets}, followed by the evaluation and discussion in Section \ref{sec:evaluation}. The paper finishes with conclusions and ideas for future research directions.

\section{Related Work}
\label{sec:related}

During the past decade there has been tremendous progress in the field of quality estimation, largely as a result of the QE shared tasks organised annually by the Workshops on Statistical Machine Translation (WMT), more recently called the Conferences on Machine Translation, since 2012. The annotated datasets these shared tasks released each year have led to the development of many open-source QE systems like QuEst \cite{specia-etal-2013-quest}, QuEst++ \cite{specia-etal-2015-multi}, deepQuest \cite{ive-etal-2018-deepquest}, and OpenKiwi \cite{kepler-etal-2019-openkiwi}. Before the neural network era, most of the quality estimation systems like QuEst \cite{specia-etal-2013-quest} and QuEst++ \cite{specia-etal-2015-multi} were heavily dependent on linguistic processing and feature engineering to train traditional machine-learning algorithms like support vector regression and randomised decision trees \cite{specia-etal-2013-quest}. Even though, they provided good results, these traditional approaches are no longer the state of the art. In recent years, neural-based QE systems have consistently topped the leader boards in WMT quality estimation shared tasks  \cite{kepler-etal-2019-openkiwi}. 

For example, the best-performing system at the WMT 2017 shared task on QE was \textsc{POSTECH}, which is purely neural and does not rely on feature engineering at all \cite{kim-etal-2017-predictor}. \textsc{POSTECH} revolves around an encoder-decoder Recurrent Neural Network (RNN) (referred to as the 'predictor'), stacked with a bidirectional RNN (the 'estimator') that produces quality estimates. In the predictor, an encoder-decoder RNN model predicts words based on their context representations and in the estimator step there is a bidirectional RNN model to produce quality estimates for words, phrases and sentences based on representations from the predictor. To be effective, \textsc{POSTECH} requires extensive predictor pre-training, which means it depends on large parallel data and is computationally intensive \cite{ive-etal-2018-deepquest}. The \textsc{POSTECH} architecture was later re-implemented in deepQuest \cite{ive-etal-2018-deepquest}. 

OpenKiwi \cite{kepler-etal-2019-openkiwi} is another open-source QE framework developed by Unbabel. It implements four different neural network architectures \textsc{QUETCH} \cite{kreutzer-etal-2015-quality}, \textsc{NuQE} \cite{martins-etal-2016-unbabels}, Predictor-Estimator \cite{kim-etal-2017-predictor} and a stacked model of those architectures. Both the \textsc{QUETCH} and \textsc{NuQE} architectures have simple neural network models that do not rely on additional parallel data, but do not perform that well. The Predictor-Estimator model is similar to the \textsc{POSTECH} architecture and relies on additional parallel data. In OpenKiwi, the best performance for sentence-level quality estimation was given by the stacked model that used the Predictor-Estimator model, meaning that the best model requires extensive predictor pre-training and relies on large parallel data and computational resources.    

In order to remove the dependency on large parallel data, which also entails the need for powerful computational resources, we propose to use crosslingual embeddings that are already fine-tuned to reflect properties between languages. We assume that by using them we will ease the burden of having complex neural network architectures. Over the last few years there has been significant work done in the area of crosslingual embeddings \cite{10.1613/jair.1.11640}. 

Since the introduction of BERT \cite{devlin2019bert}, transformer models have been used successfully for various NLP tasks such as named entity recognition \cite{devlin2019bert}, sentence classification \cite{10.1007/978-3-030-32381-3_16}, and question answering \cite{devlin2019bert}, in many cases improving the state of the art. Most of the tasks were focused on English due to the fact that most of the pre-trained transformer models were trained on English data. Although there are several multilingual models like multilingual BERT (mBERT) \cite{devlin2019bert} and multilingual DistilBERT (mDistilBERT) \cite{Sanh2019DistilBERTAD}, researchers expressed some reservations about their ability to represent all the languages \cite{pires-etal-2019-multilingual}. In addition, although mBERT and mDistilBERT showed some crosslingual characteristics, they do not perform well on crosslingual benchmarks \cite{karthikeyan2020cross}. 

XLM-RoBERTa (XML-R) was released in November 2019 \cite{conneau2019unsupervised} as an update to the XLM-100 model \cite{lample2019cross}. XLM-R takes a step back from XLM, eschewing XLM's Translation Language Modeling (TLM) objective since it requires a dataset of parallel sentences, which can be difficult to acquire. Instead, XLM-R trains RoBERTa\cite{liu2019roberta} on a huge, multilingual dataset at an enormous scale: unlabelled text in 104 languages is extracted from CommonCrawl datasets, totalling 2.5TB of text. It is trained using only RoBERTa's \cite{liu2019roberta} masked language modelling (MLM) objective. Surprisingly, this strategy provided better results in crosslingual tasks. XLM-R outperforms mBERT on a variety of crosslingual benchmarks such as crosslingual natural language inference and crosslingual question answering
 \cite{conneau2019unsupervised}. 

Both architectures proposed in \textit{\TransQuest} have been successfully applied in the monolingual semantic textual similarity tasks \cite{devlin2019bert,reimers-gurevych-2019-sentence}. When applied in monolingual experiments, both of them use monolingual transformer models like BERT \cite{devlin2019bert}, RoBERTa \cite{liu2019roberta} as the input. This inspired us to change the input in such a way that it can represent both the source and target sentences for which the quality of translation needs to be estimated, with the hope that the same architectures would also provide good results in the QE task. Our initial experiments showed that crosslingual embeddings like XLM-R provide better results than multilingual embeddings like mBERT. Therefore, in this research we explore the performance of crosslingual embeddings with simple neural network architectures for the sentence-level quality estimation task. To the best of our knowledge, state-of-the-art crosslingual contextual embeddings such as XLM-R have not been used in quality estimation before.

\section{Methodology}
\label{sec:methodology}

This section presents the methodology used to develop our quality estimation methods. 
We first describe the neural network architectures we proposed, followed by the method used to train these architectures. 

\subsection{Neural Network Architectures}
\label{sebsec:archi}

The \textit{\TransQuest} framework that is used to implement the two architectures described here relies on the XLM-R transformer model \cite{conneau2019unsupervised} to derive the representations of the input sentences. The XLM-R transformer model takes a sequence of no more than 512 tokens as input and outputs the representation of the sequence. The first token of the sequence is always \textsc{[CLS]}, which contains the special embedding to represent the whole sequence, followed by embeddings acquired for each word in the sequence. As shown below, our proposed neural network architectures can utilise both the embedding for the \textsc{[CLS]} token and the embeddings generated for each word. The output of the transformer (or transformers for \textbf{Siamese\TransQuest} described below), is fed into a simple output layer which is used to estimate the quality of translation. We describe below the way the XLM-R transformer is used and the output layer, as they are different in the two instantiations of the framework. The fact that we do not rely on a complex output layer makes training our architectures much less computational intensive than alternative solutions. The \textit{\TransQuest} framework is open-source, which means researchers can easily propose alternative architectures to the ones we present in this paper.

Both neural network architectures presented below use the pre-trained XLM-R-large model released by HuggingFace's Transformers library \cite{Wolf2019HuggingFacesTS}. The XLM-R-large model covers 104 languages \cite{conneau2019unsupervised}, making it potentially very useful to estimate the translation quality for a large number of language pairs.

\begin{figure}

  \begin{subfigure}[b]{6cm}
    \centering\includegraphics[width=5cm]{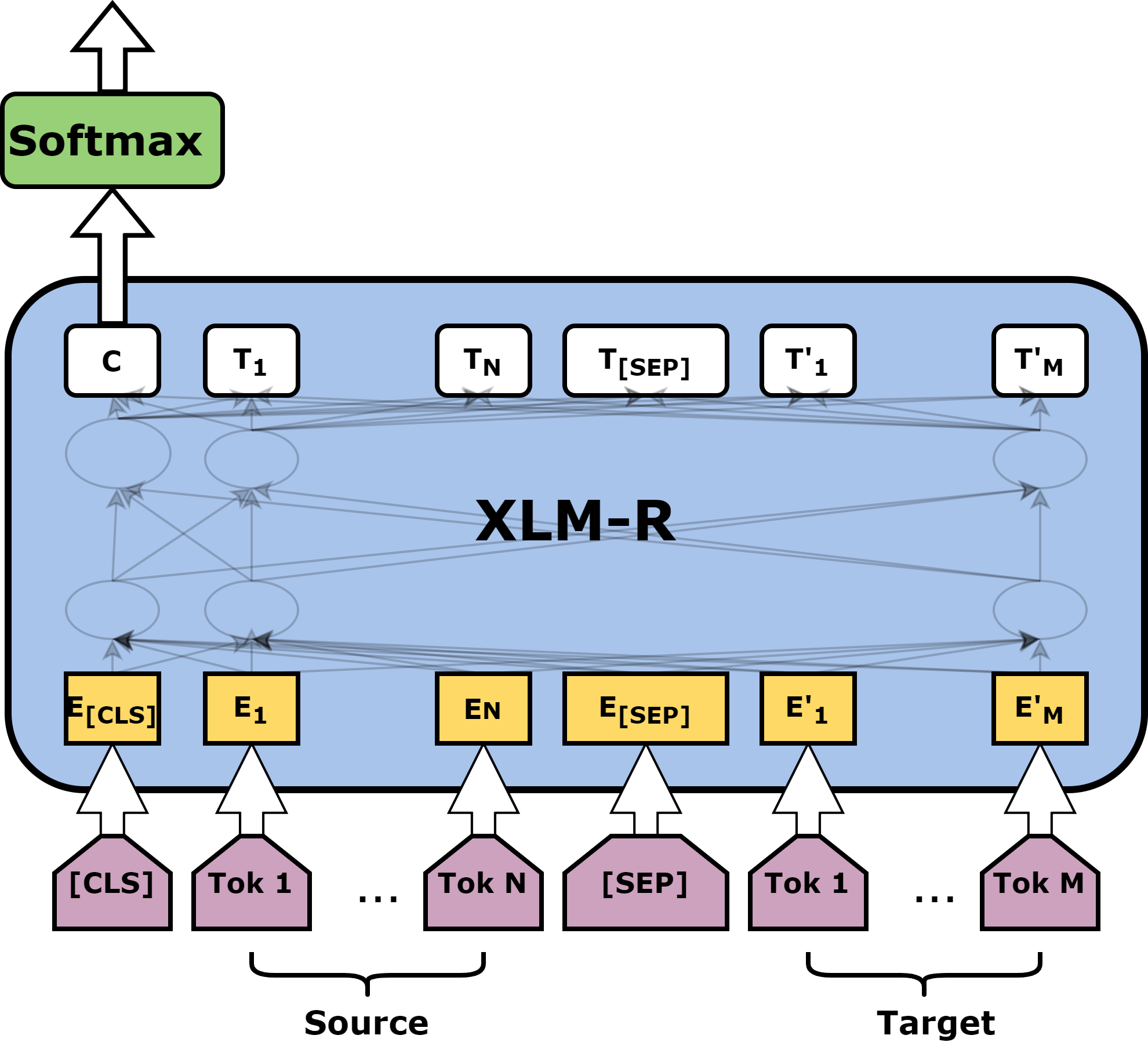}
    \caption{\textit{M\TransQuest} architecture}
    \label{fig:transquest_architecture}
  \end{subfigure}
  \begin{subfigure}[b]{6cm}
    \centering\includegraphics[width=9.5cm]{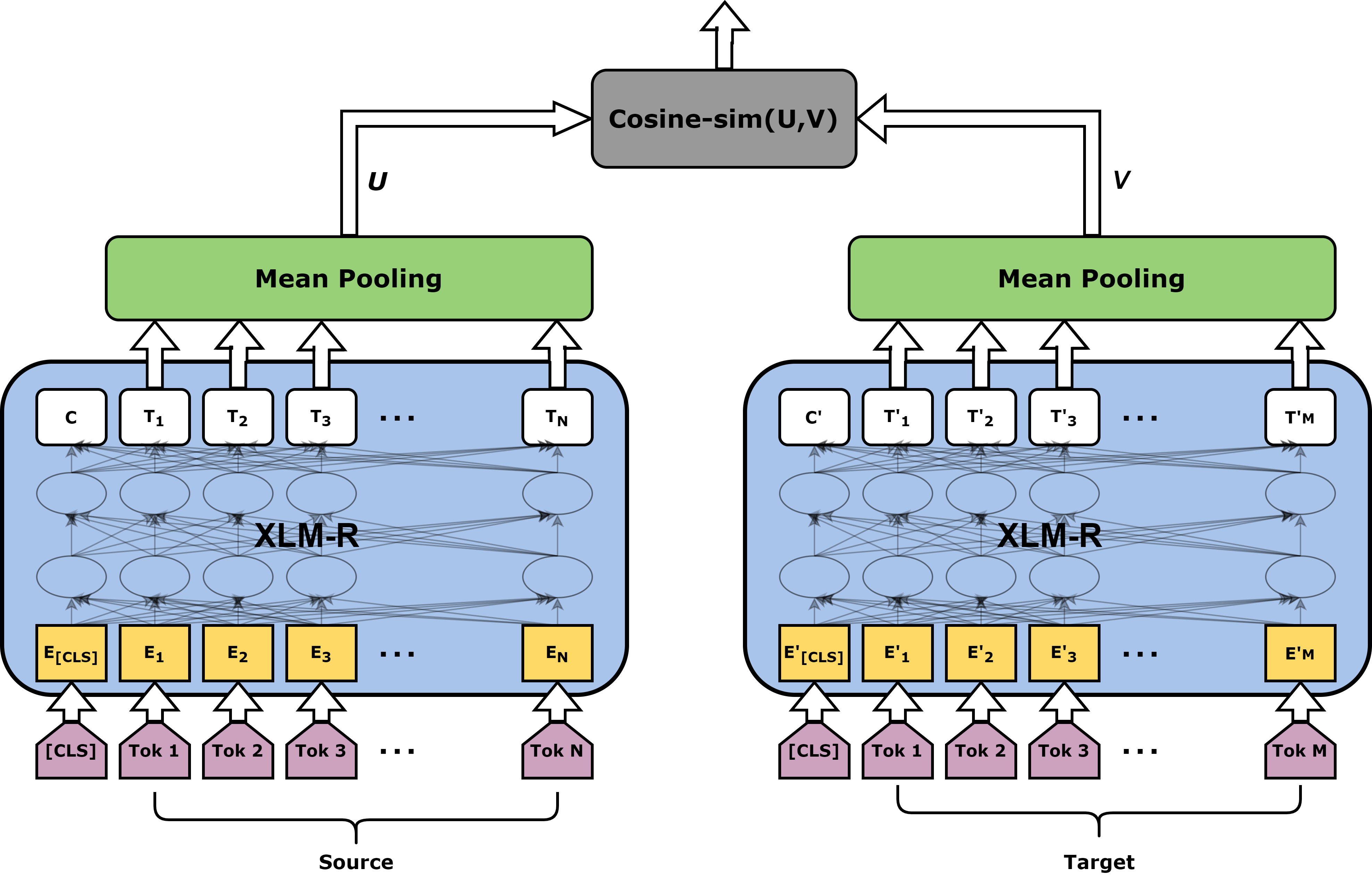}
    \caption{\textit{S\TransQuest} Architecture}
    \label{fig:siamese_transquest_architecture}
  \end{subfigure}
\caption{Two architectures of the \textit{\TransQuest} framework.}
\label{fig:architectures}
\end{figure}

\textit{\TransQuest} implements two different neural network architectures to perform sentence-level translation quality estimation which we describe below. The architectures are presented in Figure \ref{fig:architectures}.

\begin{enumerate}
  \item \textbf{Mono\TransQuest} (\textbf{M\TransQuest}): 
  The first architecture proposed uses a single XLM-R transformer model and is shown in Figure \ref{fig:transquest_architecture}. The input of this model is a concatenation of the original sentence and its translation, separated by the \textsc{[SEP]} token. We experimented with three pooling strategies for the output of the transformer model: using the output of the \textsc{[CLS]} token (\texttt{CLS}-strategy); computing the mean of all output vectors of the input words (\texttt{MEAN}-strategy); and computing a max-over-time of the output vectors of the input words (\texttt{MAX}-strategy). The output of the pooling strategy is used as the input of a softmax layer that predicts the quality score of the translation. We used mean-squared-error loss as the objective function. Early experiments we carried out demonstrated that the \texttt{CLS}-strategy leads to better results than the other two strategies for this architecture. Therefore,  we used
  the embedding of the \textsc{[CLS]} token as the input of a softmax layer.

  \item \textbf{Siamese\TransQuest} (\textbf{S\TransQuest}): The second approach proposed in this paper relies on the Siamese architecture depicted in Figure \ref{fig:siamese_transquest_architecture} which has shown promising results in monolingual semantic textual similarity tasks \cite{reimers-gurevych-2019-sentence,ranasinghe-etal-2019-semantic}. In this case, we feed the original text and the translation into two separate XLM-R transformer models. Similar to the previous architecture we used the same three pooling strategies for the outputs of the transformer models. We then calculated the cosine similarity between the two outputs of the pooling strategy. We used mean-squared-error loss as the objective function. In initial experiments we carried out with this architecture, the \texttt{MEAN}-strategy showed better results than the other two strategies. For this reason, we used the \texttt{MEAN}-strategy for our experiments. Therefore, cosine similarity is calculated between the the mean of all output vectors of the input words produced by each transformer. 
  %As shown in Figure \ref{fig:siamese_transquest_architecture}, cosine similarity between the two outputs of the mean pooling layers was calculated to reflect the quality of the translation.
 
\end{enumerate}

 \subsection{Training Details}
 \label{subsec:training}

We used the same set of configurations for all the language pairs evaluated in this paper in order to ensure consistency between all the languages. This also provides a good starting configuration for researchers who intend to use \TransQuest on a new language pair. In both architectures we used a batch-size of eight, Adam optimiser with learning rate $2\mathrm{e}{-5}$, and a linear learning rate warm-up over 10\% of the training data. During the training process, the parameters of XLM-R model, as well as the parameters of the subsequent layers, were updated. The models were trained using only training data. Furthermore, they were evaluated while training using an evaluation set that had one fifth of the rows in training data. We performed early stopping if the evaluation loss did not improve over ten evaluation steps. All the models were trained for three epochs. For some of the experiments, we used an Nvidia Tesla K80 GPU, whilst for others we used an Nvidia Tesla T4 GPU. This was purely based on the availability of the hardware and it was not a methodological decision.

\section{Dataset}
\label{sec:datasets}

We used the architectures described above to predict two standard measures that express the quality of a translation: Human-mediated Translation Edit Rate (HTER) and Direct Assessment (DA).  All the datasets that we used are publicly available and were released in WMT quality estimation tasks in recent years \cite{specia-etal-2018-findings,fonseca-etal-2019-findings,specia2020findings}. This was done to ensure replicability of our experiments and to allow us to compare our results with the state of the art. In the reminder of the section we provide more details about the datasets used. 

\subsection{Predicting HTER}
The performance of QE systems has typically been assessed using the semiautomatic HTER (Human-mediated Translation Edit Rate). HTER is an edit-distance-based measure which captures the distance between the automatic translation and a reference translation in terms of the number of modifications required to transform one into another. In light of this, a QE system should be able to predict the percentage of edits required in the translation. We used several language pairs for which HTER information was available: English-Chinese (En-Zh), English-Czech (En-Cs), English-German (En-De), English-Russian (En-Ru), English-Latvian (En-Lv) and German-English (De-En). The texts are from a variety of domains and the translations were produced using both neural and statistical machine translation systems. More details about these datasets can be found in Table \ref{tab:hter_data} and in \cite{specia-etal-2018-findings,fonseca-etal-2019-findings}.

\begin{table}[t]
\begin{center}
\begin{tabular}{ |c|c|c|c|c| } 
 \hline
 \textbf{Language Pair} & \textbf{Source} & \textbf{MT system} & \textbf{Competition} &  \textbf{train, dev, test size} \\ 
  \hline
   De-En & Pharmaceutical & Phrase-based SMT & WMT 2018 & 25,963, 1,000, 1,000  \\
   \hline
  En-Zh & Wiki & fairseq based NMT & WMT 2020 & 7,000, 1,000, 1,000 \\
  \hline
  En-Cs & IT & Phrase-based SMT & WMT 2018 & 40,254, 1,000, 1,000 \\
 \hline
  En-De & IT & fairseq based NMT & WMT 2019 & 13,442, 1,000, 1,000 \\
  \hline
  En-De & IT & Phrase-based SMT & WMT 2018 & 26,273, 1,000, 1,000 \\
  \hline
  En-Ru & Tech & Online NMT & WMT 2019 & 15,089, 1,000, 1,000 \\
  \hline
  En-Lv & Pharmaceutical & Attention-based NMT & WMT 2018 & 12,936, 1,000, 1,000 \\
  \hline
  En-Lv & Pharmaceutical & Phrase-based SMT & WMT 2018 & 11,251, 1,000, 1,000 \\
  \hline
\end{tabular}
\end{center}
\caption{Information about language pairs used to predict HTER. The \textbf{Language Pair} column shows the language pairs we used in ISO 639-1 codes\protect\footnotemark. \textbf{Source} expresses the domain of the sentence and \textbf{MT system} is the Machine Translation system used to translate the sentences. 
In that column NMT indicates Neural Machine Translation and SMT indicates Statistical Machine Translation. 
\textbf{Competition} shows the quality estimation competition in which the data was released and the last column indicates the number of instances the train, development and test dataset had in each language pair respectively.} 
\label{tab:hter_data}
\end{table}

\footnotetext{Language codes are available in ISO 639-1 Registration Authority Website Online - \url{ https://www.loc.gov/standards/iso639-2/php/code_list.php}}

\subsection{Predicting DA}
Even though HTER has been typically used to assess quality in machine translations, the reliability of this metric for assessing the performance of quality estimation systems has been questioned by researchers \cite{graham-etal-2016-glitters}. The current practice in MT evaluation is the so-called Direct Assessment (DA) of MT quality \cite{graham_baldwin_moffat_zobel_2017}, where raters evaluate the machine translation on a continuous 1-100 scale. This method has been shown to improve the reproducibility of manual evaluation and to provide a more reliable gold standard for automatic evaluation metrics \cite{graham-etal-2015-accurate}. 

We used a recently created dataset to predict DA in machine translations which was released for the WMT 2020 quality estimation shared task 1 \cite{specia2020findings}. The dataset is composed of data extracted from Wikipedia for six language pairs, consisting of high-resource English-German (En-De) and English-Chinese (En-Zh), medium-resource Romanian-English (Ro-En) and Estonian-English (Et-En), and low-resource Sinhala-English (Si-En) and Nepalese-English (Ne-En), as well as a Russian-English (En-Ru) dataset which combines articles from Wikipedia and Reddit \cite{fomicheva2020unsupervised}. These datasets have been collected by translating sentences sampled from source-language articles using state-of-the-art NMT models built using the fairseq toolkit \cite{ott-etal-2019-fairseq} and annotated with DA scores by professional translators. Each translation was rated with a score from 0-100 according to the perceived translation quality by at least three translators \cite{specia2020findings}. The DA scores were standardised using the z-score. The quality estimation systems evaluated on these datasets have to predict the mean DA z-scores of test sentence pairs. Each language pair has 7,000 sentence pairs in the training set, 1,000 sentence pairs in the development set and another 1,000 sentence pairs in the testing set.

\section{Evaluation and discussion}
\label{sec:evaluation}

This section presents the evaluation results of our architectures on the datasets described in the previous section in a variety of settings. We first evaluate them in a single language pair setting (Section \ref{subsec:supervised}), which is essentially the setting employed in the WMT shared tasks. We then evaluate in a setting where we combine datasets in several languages for training (Section \ref{subsec:multi}). We conclude the section with an evaluation of a transfer learning setting (Section \ref{subsec:transfer}). 

In order to better understand the performance of our approach, we compare our results with the baselines reported by the WMT2018-2020 organisers. Two baselines were used: OpenKiwi \cite{kepler-etal-2019-openkiwi} and QuEst++ \cite{specia-etal-2015-multi}.  Row \textbf{IV} of Tables \ref{tab:hter_prediction} and \ref{tab:results:direct_assesement} present the results for these baselines. For some language pairs in the 2018 WMT quality estimation shared task the organisers did not report the scores obtained using OpenKiwi. These cases are marked with \textbf{NR} in Table \ref{tab:hter_prediction}. Row \textbf{IV} of both tables also includes the results of the best system from WMT2018 to WMT2020 for each setting. Additionally,  Row \textbf{IV} of Table \ref{tab:results:direct_assesement} shows the results of the TransQuest's submission to WMT 2020 QE Task 1, which was the winning solution is all the languages \cite{ranasinghe-etal-2020-transquest}.

The evaluation metric used was the Pearson correlation ($r$) between the predictions and the gold standard from the test set, which is the most commonly used evaluation metric in recent WMT quality estimation shared tasks \cite{specia-etal-2018-findings,fonseca-etal-2019-findings,specia2020findings}.

% For the language pairs for which the annotated test data is not yet available to download (the language pairs from WMT 2020) we obtained the values for the Pearson correlation using CodaLab, the hosting platform of the WMT 2020 QE shared task. The only exception is the results reported in Section \ref{subsec:transfer} where the number of submissions to CodaLab required to obtain all the necessary results would have exceeded our allowance. For this reason, we evaluated the development dataset using our evaluation script. As a sanity check, we used our evaluation script on the WMT 2018 data to ensure it produces the same results as CodaLab.

\renewcommand{\arraystretch}{1.2}
\begin{table*}[t]
\begin{center}
\small
% \footnotesize
\begin{tabular}{l l  c c c c c c c c} 
%\hline
\toprule
& & \multicolumn{4}{c}{\bf Mid-resource} & \multicolumn{4}{c}{\bf High-resource}\\\cmidrule(r){3-6}\cmidrule(r){7-10}
 &{\bf Method} & \makecell{En-Cs \\ SMT} & \makecell{ En-Ru \\ NMT} & \makecell{En-Lv \\ SMT} & \makecell{En-Lv \\ NMT} & \makecell{De-En \\ SMT} & \makecell{En-Zh \\ NMT} & \makecell{En-De \\ SMT } & \makecell{En-De \\ NMT} \\
\midrule
\multirow{2}{*}{\bf I} & M\TransQuest & \textbf{0.7207} & \textbf{0.7126} & 0.6592 & 0.7394 & \textbf{0.7939} & 0.6119 & 0.7137 & \textbf{0.5994}\\
& S\TransQuest & 0.6853 & 0.6723 & 0.6320 & 0.7183 & 0.7524 & 0.5821& 0.6992 & 0.5875 \\
\midrule
\multirow{2}{*}{\bf II} & M\TransQuest *-En$|$En-* & 0.7168 & 0.7046  & \textbf{0.7181} & \textbf{0.7482} & 0.7939 & 0.6101 & 0.7355 & 0.5992 \\
& S\TransQuest *-En$|$En-* & 0.6663 & 0.6701 & 0.6533 & 0.7192 & 0.7524 & 0.5721 & 0.7000 & 0.5793 \\
\midrule
\multirow{2}{*}{\bf III} & M\TransQuest-m & 0.7111 & 0.7012 & 0.7141 & 0.7450 & 0.7878 & 0.6092 & 0.7300 & 0.5982 \\
& S\TransQuest-m & 0.6561 & 0.6614 & 0.6621 & 0.7202 & 0.7369 & 0.5612 & 0.7015 & 0.5771 \\
%\midrule
\midrule
\multirow{3}{*}{\bf IV} & Quest ++ & 0.3943 & 0.2601 & 0.3528 & 0.4435 & 0.3323 & NR & 0.3653 & NR \\
& OpenKiwi & NR & 0.5923 & NR & NR & NR & 0.5058 & 0.7108 & 0.4001 \\
& Best system & 0.6918 & 0.5923 & 0.6188 & 0.6819 & 0.7888 & \textbf{0.6641} & \textbf{0.7397} &  0.5718 \\
\midrule
\multirow{1}{*}{\bf V} & mBERT & 0.6423 & 0.6354 & 0.5772 & 0.6531 & 0.7005 & 0.5483 & 0.6239 & 0.5002 \\
\bottomrule
%\bottomrule
\end{tabular}
\end{center}
\caption{Pearson ($r$) correlation between \textit{\TransQuest} algorithm predictions and human post-editing effort. Best results for each language by any method are marked in bold. Rows I, II and III indicate the different evaluation settings, explained in Sections \ref{subsec:supervised}, \ref{subsec:multi} and \ref{subsec:transfer}. Row IV shows the results of the state-of-the-art methods and the best system submitted for the language pair in that competition. \textbf{NR} implies that a particular result 
% from a state-of-the-art method 
was \textit{not reported} by the organisers. Row V presents the results of the multilingual BERT (mBERT) model in MonoTransQuest Architecture.} 
\label{tab:hter_prediction}
\end{table*}

\renewcommand{\arraystretch}{1.2}
\begin{table*}[t]
\begin{center}
\small
% \footnotesize
\begin{tabular}{l l  c c c c c c c} 
%\hline
\toprule
& & \multicolumn{2}{c}{\bf Low-resource} & \multicolumn{3}{c}{\bf Mid-resource} & \multicolumn{2}{c}{\bf High-resource}\\\cmidrule(r){3-4}\cmidrule(lr){5-7}\cmidrule(l){8-9}
 &{\bf Method} & Si-En & Ne-En & Et-En & Ro-En & Ru-En & En-De & En-Zh\\
\midrule
\multirow{2}{*}{\bf I} & M\TransQuest & 0.6525 & \textbf{0.7914} & 0.7748 & \textbf{0.8982} & 0.7734 & \textbf{0.4669} & \textbf{0.4779} \\
& S\TransQuest & 0.5957 & 0.7081 & 0.6804 & 0.8501 & 0.7126 & 0.3992 & 0.4067 \\
\midrule
\multirow{2}{*}{\bf II} & M\TransQuest *-En$|$En-* & \textbf{0.6528} & 0.7824 & \textbf{0.7827} & 0.8868 & \textbf{0.7821} & 0.4518 & 0.4334 \\
& S\TransQuest *-En$|$En-* & 0.5968 & 0.6992 & 0.6921 & 0.8432 & 0.7152  & 0.3621 & 0.3812 \\
\midrule
\multirow{2}{*}{\bf III} & M\TransQuest-m &  0.6526 & 0.7581 & 0.7574 & 0.8856  &  0.7521 &  0.4420 &  0.4646 \\
& S\TransQuest-m & 0.5970 & 0.6980 & 0.6934 & 0.8426 & 0.6945 & 0.3832 &  0.3900 \\
\midrule
\multirow{2}{*}{\bf IV} & OpenKiwi & 0.3737 & 0.3860 & 0.4770 & 0.6845 & 0.5479 & 0.1455 & 0.1902 \\
& \TransQuest @WMT2020 & 0.6849 & 0.8222 & 0.8240 & 0.9082 & 0.8082 & 0.5539 & 0.5373 \\
\midrule
\multirow{1}{*}{\bf V} & mBERT & NS & 0.6452 & 0.6231 & 0.8351 & 0.6661 & 0.3765 & 0.3982 \\

\bottomrule
\end{tabular}
\end{center}
\caption{Pearson ($r$) correlation between \textit{\TransQuest} algorithm predictions and human DA judgments. Best results for each language (any method) are marked in bold. Rows I, II and III indicate the different settings of \textit{\TransQuest}, explained in Sections \ref{subsec:supervised}-\ref{subsec:transfer}. OpenKiwi baseline results are in Row IV. Row IV also shows the results for TransQuest's submission to WMT 2020 QE Task 1 \protect\cite{ranasinghe-etal-2020-transquest} which was also the winning solution. Row V presents the results of the multilingual BERT (mBERT) model in MonoTransQuest Architecture. \textbf{NS} implies that the non-English language in the language pair is not supported by mBERT. }
\label{tab:results:direct_assesement}
\end{table*}

\subsection{Supervised Single Language Pair Quality Estimation}
\label{subsec:supervised}

The first evaluation we carried out was the \textit{supervised single language pair} evaluation where we used the training set of each language to build a quality estimation model and we evaluated it on a testing set from the same language. 
This replicates the standard QE evaluation carried out in the WMT shared tasks.
The results for each language in \textit{supervised} settings are shown in row I of Tables \ref{tab:hter_prediction} and \ref{tab:results:direct_assesement}. The results indicate that both architectures proposed in \textit{\TransQuest} outperform the baselines in all the language pairs of both aspects in quality estimation, and also outperform the best systems from previous competitions. From the two architectures, \textit{M\TransQuest} performs slightly better than \textit{S\TransQuest}. 

In the HTER aspect of quality estimation, as shown in Table \ref{tab:hter_prediction}, \textit{M\TransQuest} gains $\approx$ 0.1-0.2 Pearson correlation boost over OpenKiwi in most language pairs. However, OpenKiwi comes very close to \textit{M\TransQuest} in En-De SMT.
In the language pairs where OpenKiwi results are not available \textit{M\TransQuest} gains $\approx$ 0.3-0.4 Pearson correlation boost over QuEst++ in all language pairs for both NMT and SMT. Table \ref{tab:hter_prediction}  also gives the results of the best system submitted for a particular language pair. It is worth noting that for the training setting described in this section, the \textit{\TransQuest} results surpass the best system in all the language pairs with the exception of the En-De SMT and En-Zh NMT datasets. 

As shown in Table \ref{tab:results:direct_assesement}, in the DA aspect of quality estimation, \textit{M\TransQuest}  gained $\approx$ 0.2-0.3 Pearson correlation boost over OpenKiwi in all the language pairs. Additionally, \textit{M\TransQuest} achieves $\approx$ 0.4 Pearson correlation boost over OpenKiwi in the low-resource language pair Ne-En. Furthermore, TransQuest participated in WMT 2020 quality estimation shared task 1 and it was the winning solution in all the language pairs. To achieve this restult, TransQuest was fine-tuned with self-ensemble and data augmentation to achieve the first place. We do not describe here the fine tuning approaches since they are task specific, but more details can be found in \cite{ranasinghe-etal-2020-transquest}.

Additionally, row V in both Tables \ref{tab:hter_prediction} and \ref{tab:results:direct_assesement} shows the results of multilingual BERT (mBERT) in MonoTransQuest architecture. We used the same settings similar to XLM-R. The results show that XLM-R model outperforms the mBERT model in all the language pairs of both aspects in quality estimation and we can safely assume that the cross lingual nature of the XLM-R transformers had a clear impact to the results.

\subsection{Supervised Multi-Language Pair Quality Estimation}
\label{subsec:multi}

Most of the available open-source quality estimation frameworks require maintaining separate machine learning models for each language. This can be very challenging in a practical environment where the systems have to work with 10-20 language pairs. Furthermore, pre-trained neural quality estimation models are large. In a commercial environment, where the quality estimation systems need to do inference on several language pairs, loading all of the pre-trained models from all language pairs would require a lot of Random Access Memory (RAM) space and result in a huge cost.    

Therefore, with \textit{\TransQuest} we propose a single model that can perform quality estimation on several language pairs. We propose two training strategies for \textit{supervised multi language pair} settings:

\begin{enumerate}
  \item We separate the language pairs into two groups. One group contains all the language pairs where the source language is English and in the other group the target is always English. We represent the former with En-$*$, and the latter with $*$-En. We train both architectures by concatenating training sets in all the language pairs in a particular group. To ease the comparison of the results, we evaluate the model separately on each language pair. We do this process for both aspects in quality estimation. The results are shown in row II in Tables \ref{tab:hter_prediction} and  \ref{tab:results:direct_assesement}.
  
  \item We concatenate training data from all the language pairs, without considering the direction of the translation, and build a single model for all language pairs. We refer to these models by \textit{M\TransQuest-m} and \textit{S\TransQuest-m}. Similarly to the first \textit{multi-language pair} training strategy, we evaluate the model separately on each language pair 
  %to ease the comparison of the results and repeat this process 
  and for both aspects of quality estimation. The results are shown in row III in Tables \ref{tab:hter_prediction} and \ref{tab:results:direct_assesement}. 
  
\end{enumerate}

As depicted in Tables \ref{tab:hter_prediction} and \ref{tab:results:direct_assesement}, the multi-language pair experiments yielded very competitive results.
In fact, for some language pairs the multi-language pair model performed better than the model that was trained solely on that particular pair. When predicting HTER, the multi-language pair model performed better in En-Lv for both SMT and NMT, and in En-De for SMT while performing on par in En-De for NMT. In predicting DA, the multi-language pair model performed better in Et-En, Ru-En and Si-En. In addition, with the exceptions of the En-De SMT and En-Zh NMT setting, the models that consider the direction of the language pairs are better than the best systems submitted to previous editions of WMT.

Throughout our experiments we noted that the \TransQuest models built with the direction of the language pairs in mind performed slightly better than the \TransQuest models trained without considering the language pair direction. It should be noted that none of the multi-language pair models' Pearson correlation decreased by more than 0.03\% in any language pair for either \TransQuest architecture. Similar to \textit{supervised single language pair} experiments, M\TransQuest architecture performed better than S\TransQuest architecture in all the languages in both aspects of quality estimation. 

The size of the pre-trained \textit{\TransQuest} models on a single language pair was $\approx$ 2GB. The pre-trained models for multiple language pairs in this section did not exceed more than 2.1 GB. Therefore, we present multi-language pair pre-trained models as a solution for environments that are on tight resources and seek to conduct quality estimation on multiple language pairs.

\subsection{Transfer Learning based Quality Estimation}
\label{subsec:transfer}

The biggest challenge in building supervised quality estimation models is not having enough annotated data \cite{fomicheva2020unsupervised}, especially for low-resourced languages. We explore the possibility of performing transfer learning on low-resource languages using the models trained on better-resourced languages. As the low-resource language pairs were only available in the DA aspect, we conducted this experiment only in the DA aspect in quality estimation. All of the low-resource language pairs in the DA aspect had English as the target language. Considering the positive impact that the direction of the language pair had on Pearson correlation in the previous experiment, we decided to consider only the language pairs with English as the target language. This left us only with mid-resource language pairs for training. We were also aware that the {\TransQuest} models had relatively low Pearson correlations with high-resource language pairs in DA.

\begin{enumerate}
\item We build a single model for each architecture using all the training data available for mid-resource language pairs: Et-En, Ro-En and Ru-En. We refer to this as \textit{\TransQuest-Mid}. 

\item When we train a \textit{\TransQuest} model for a low-resource language pair, we initiate the model weights from \textit{\TransQuest-Mid} and start training. To see whether it is possible to get compatible results even with fewer training instances, we conduct the experiments for 0 (unsupervised), 100, 200, 300 and up to 1,000 training sentence pairs. We do this for Si-En and Ne-En. Depending on the architecture we use, we refer to this model as \textit{M\TransQuest TL} or \textit{S\TransQuest TL}.

\item In order to evaluate the effect of transfer learning we conduct the same experiment in step 2, but we train the model from scratch. Depending on the architecture we use, we refer to this model as \textit{M\TransQuest Scratch} or \textit{S\TransQuest Scratch}. 
%without initiating the model weights from \textit{\TransQuest-Mid}. 

\end{enumerate}

\begin{figure}
\centering
  \begin{subfigure}[b]{7cm}
    \centering\includegraphics[width=7cm]{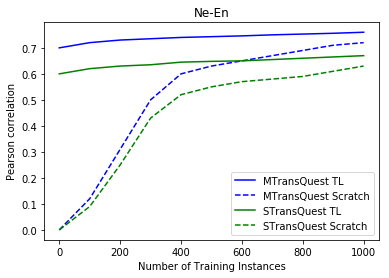}
    \caption{Ne-En Results}
    \label{fig:ne_en_results}
  \end{subfigure}
  \begin{subfigure}[b]{7cm}
    \centering\includegraphics[width=7cm]{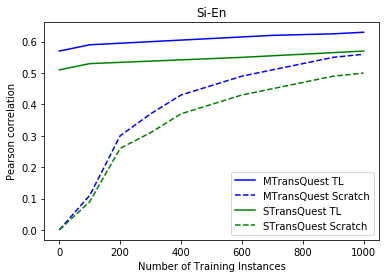}
    \caption{Si-En Results}
    \label{fig:si_en_Results}
  \end{subfigure}
\caption{Transfer learning impact of \textit{\TransQuest} Architectures. \textbf{\textit{M\TransQuest} Scratch} and \textbf{\textit{S\TransQuest} Scratch} indicates that the model was trained from scratch while \textbf{\textit{M\TransQuest} TL} and \textbf{\textit{S\TransQuest} TL} indicates that models followed the transfer learning strategy.}
\label{fig:tl_results}
\end{figure}

As shown in Figure \ref{fig:tl_results}, the transfer learning strategy significantly impacts the results. In Ne-En, with only 100 training instances, training \textit{M\TransQuest scratch} achieves only 0.1242 Pearson correlation between the predictions and gold labels of the test set. However, in Ne-En with only 100 training instances, training M\textit{\TransQuest} using the transfer learning strategy achieves 0.7417 Pearson correlation, which is close to the best result obtained with the \textit{M\TransQuest} architecture for Ne-En after training with 7,000 instances (0.7914). When the number of training instances grows, the results from the \textit{\TransQuest} models trained with the transfer learning strategy and the results from the \textit{\TransQuest} models trained from scratch converge. A similar pattern, but with a lower Pearson correlation, can be seen with \textit{S\TransQuest}. Similar results can also be observed for the other low-resource language pair, Si-En. Therefore, it is safe to conclude that \textit{\TransQuest} with the transfer learning strategy can be hugely beneficial to low-resource language pairs in quality estimation where annotated training instances are scarce.  

\section{Conclusions}

In this paper we introduced \textit{\TransQuest}, a new open source framework for quality estimation based on cross-lingual transformers. \textit{\TransQuest} is implemented in PyTorch and supports training of sentence-level quality estimation systems on new data. It outperforms other open-source tools on both aspects of sentence-level quality estimation and yields new state-of-the-art quality estimation results. As far as we know, it is the first time that an open-source QE framework has been tested on both aspects of quality estimation. Furthermore, it is the first time that a QE system explores multi-language pair models and transfer learning on low-resource language pairs. Unlike many other open-source neural QE frameworks, \TransQuest does not use parallel data  and hence does not require similar computational resources.

We propose two architectures: \textit{M\TransQuest} and \textit{S\TransQuest}, neither of which have been previously explored in QE tasks. The two architectures have a trade-off between accuracy and efficiency. On an Nvidia Tesla K80 GPU, \textit{M\TransQuest} takes 4,480s on average to train on 7,000 instances, while \textit{S\TransQuest} takes only 3,900s on average for the same experiment. On the same GPU, \textit{M\TransQuest} takes 35s on average to perform inference on 1,000 instances which takes \textit{S\TransQuest} only 16s to do so. Therefore we recommend using \textit{M\TransQuest} where accuracy is valued over efficiency, and S\TransQuest where efficiency is prioritised above accuracy. Since there is a growing interest\footnote{Several workshops like $EMC^2$ (Workshop on Energy Efficient Machine Learning and Cognitive Computing), SustaiNLP 2020 (Workshop on Simple and Efficient Natural Language Processing) has been organised on this aspect.} in the NLP community for energy efficient machine learning models, we decided to support both architectures in the TransQuest Framework.   

In the future, we plan to expand \TransQuest with more neural network architectures and more models for different levels of quality estimation such as word-level and document-level. In the sentence-level, we plan to perform transfer learning on language pairs that do not include English at all. We also hope to conduct unsupervised experiments on low-resource language pairs. 

% \section*{Acknowledgement}

% We would like to thank WMT QE shared task organisers for making the datasets used in this paper publicly available. We further thank the anonymous COLING reviewers for providing us with valuable feedback to improve this paper.

% We are also grateful for our colleagues in RGCL who provided their support with proofreading and providing valuable feedback to this paper. 

\bibliographystyle{coling}
\bibliography{coling2020}

\end{document}